\documentclass[runningheads]{llncs}
\usepackage[T1]{fontenc}
\usepackage{graphicx}
\usepackage{amsmath,amssymb} 

\usepackage{epsfig}
\usepackage{bbold}
\usepackage{dsfont}
\usepackage{enumerate}
\usepackage{enumitem}
\usepackage[toc,page]{appendix}

\usepackage{xcolor}
\usepackage{bm}
\usepackage{isomath}
\usepackage{arydshln}
\usepackage{stmaryrd}

\usepackage{amsfonts}
\usepackage{dsfont}
\usepackage{bm}
\usepackage{mathtools}

\usepackage{dblfloatfix}
\usepackage{pbox}
\usepackage{capt-of}
\usepackage[normalem]{ulem}
\usepackage{multirow}
\usepackage{colortbl}
\usepackage{mathtools}
\usepackage{array}
\usepackage{indentfirst}
\usepackage{booktabs}
\usepackage{xspace}
\usepackage{colortbl}

\usepackage[pagebackref,breaklinks,colorlinks,bookmarks=false,urlcolor=black
,citecolor={green!60!black}]{hyperref}

\usepackage{orcidlink}

\definecolor{cGreen}{RGB}{0,255,0}
\definecolor{LightCyan}{rgb}{0.8,1,0.8}

\newcommand{\comment}[1]{}

\makeatletter
\renewcommand{\paragraph}{%
  \@startsection{paragraph}{4}%
  {\z@}{0.75ex \@plus 1ex \@minus 0.9ex}{-0.3em}%
  {\normalfont\normalsize\bfseries}%
}
\makeatother

\renewcommand\vec[1]{\ensuremath\boldsymbol{#1}}
\renewcommand\cdots{...}

\newcommand{\tF}{\vec{\mathcal{F}}}

\newcommand{\vy}{\mathbf{y}}

\newcommand{\mX}{\mathbf{X}}
\newcommand{\vx}{\mathbf{x}}

\newcommand{\mbr}[1]{\mathbb{R}^{#1}}

\newcommand{\idx}[1]{\mathcal{I}_{#1}}

\newcommand{\tR}{\vec{\mathcal{R}}}

\newcommand{\vphi}{\boldsymbol{\phi}}

\newcommand{\mPsi}{\vec{\Psi}}

\newcommand{\set}[1]{\left\{#1\right\}}

\DeclareMathOperator*{\argmin}{arg\,min}

\newcommand{\tM}{\vec{\mathcal{M}}}
\newcommand{\tD}{\vec{\mathcal{D}}}
\newcommand{\tC}{\vec{\mathcal{C}}}

\def\eg{\emph{e.g.}}

\newcommand{\mPhi}{\boldsymbol{\Phi}}

\newcommand{\stkout}[1]{{\ifmmode\text{\sout{\ensuremath{#1}}}\else\sout{#1}\fi}}

\makeatletter
\DeclareRobustCommand\onedot{\futurelet\@let@token\bmv@onedotaux}
\def\bmv@onedotaux{\ifx\@let@token.\else.\null\fi\xspace}
\def\eg{\emph{e.g}\onedot} 
\def\ie{\emph{i.e}\onedot} 
 
\def\etc{\emph{etc}\onedot} \def\vs{\emph{vs}\onedot}
\def\wrt{w.r.t\onedot} 
\def\etal{\emph{et al}\onedot}

\makeatother

\begin{document}
\title{Improving Few-shot Learning by Spatially-aware Matching and CrossTransformer}
\titlerunning{Improving FSL by SM and CrossTransformer.}
\author{Hongguang Zhang\inst{1}\orcidlink{0000-0003-0734-0078} \and
Philip H. S. Torr\inst{4} \and
Piotr Koniusz\inst{2,3}\orcidlink{0000-0002-6340-5289}}
\authorrunning{Zhang \etal}
%
\institute{Systems Engineering Institute, AMS \\ \email{zhang.hongguang@outlook.com} 
\and Data61/CSIRO \\ \email{piotr.koniusz@data61.csiro.au}  
\and Australian National University \\
\and Oxford University}
\maketitle              
\begin{abstract}
Current few-shot learning models capture  visual object relations in the so-called meta-learning setting under a fixed-resolution input. However, such models have a limited generalization ability under the scale and location mismatch between objects, as only  few samples from target classes are provided. Therefore, the lack of a mechanism to match the scale and location between pairs of compared images leads to the performance degradation. The importance of image contents varies across coarse-to-fine scales depending on the object and its class label, \eg, generic objects and scenes rely on their global appearance while fine-grained objects rely more on their localized visual patterns. In this paper, we study the impact of scale and location mismatch in the few-shot learning scenario, and propose a novel Spatially-aware Matching (SM) scheme to effectively perform matching across multiple  scales and locations, and learn image relations by giving the highest weights to  the best matching pairs.  The SM is trained to activate the most related locations and scales between support and query data. We apply and evaluate SM on various few-shot learning models and backbones for comprehensive evaluations. Furthermore, we leverage an auxiliary self-supervisory discriminator to train/predict the spatial- and scale-level index of feature vectors we use. Finally, we develop a novel transformer-based pipeline to exploit self- and cross-attention in a spatially-aware matching process. Our proposed design is orthogonal to the choice of backbone and/or comparator.
\end{abstract}
%
%
\section{Introduction}

CNNs are the backbone of object categorization, scene classification and fine-grained image recognition models but they require large amounts of labeled data. In contrast, humans enjoy the ability to learn and recognize novel objects and complex visual concepts from very few samples, which highlights the superiority of biological vision over CNNs. Inspired by the brain ability to learn in the few-samples regime, researchers study the so-called problem of few-shot learning for which networks are adapted by the use of only few training samples. Several proposed  relation-learning deep networks  \cite{vinyals2016matching,snell2017prototypical,sung2017learning,zhang2019power}  can be viewed as performing a variant of metric learning, which they are fail to address the scale- and location-mismatch between support and query samples as shown in Figure \ref{fig:scale_mismatch}.
We follow such models and focus on studying how to capture the most discriminative object scales and locations to perform accurate matching between the so-called query and support representations.

A typical relational few-shot learning pipeline consists of (i) feature encoder (backbone), (ii) pooling operator which aggregates feature vectors of query and support images of an episode followed by forming a relation descriptor, and (iii) comparator (base learner). In this paper, we investigate how to efficiently apply spatially-aware matching between query and support images across different locations and scales. To this end, we propose a Spatially-aware Matching (SM) scheme which scores the compared locations and scales. The scores can be regularized to induce sparsity and used to re-weight similarity learning loss operating on $\{0,1\}$ labels (different/same class label). Note that our SM is orthogonal to the choice of baseline, therefore it is applicable to many existing few-shot learning pipelines.

As the spatial size (height and width) of convolutional representations vary, pooling is required before feeding the representations into the comparator. We compare several pooling strategies, \ie, average, max and second-order pooling (used in object, texture and action recognition, fine-grained recognition, and few-shot learning \cite{porikli2006tracker,guo2013action,carreira_secord,koniusz2017higher,koniusz2018deeper,zhang2019power,wertheimer2019few}) which captures covariance of features per region/scale. Second-order Similarity Network (SoSN) \cite{zhang2019power} is the first work which validates the usefulness of autocorrelation representations in few-shot learning. In this paper, we employ second-order pooling as it is permutation-invariant \wrt~the spatial location of aggregated vectors while capturing second-order statistics which are more informative than typical average-pooled first-order features. 
As second-order pooling can aggregate any number of feature vectors  into a fixed-size representation, it is useful in describing regions of varying size for spatially-aware matching. 

\begin{figure*}[t]
    \centering
    \includegraphics[width=\linewidth]{images/mismatch-case.pdf}
    \caption{Scale mismatch in \textit{mini}-ImageNet. Top row: support samples randomly selected from  episodes. Remaining rows: failure queries are marked by red boxes. We estimated that  $\sim$30\% mismatches are due to the object scale mismatch, which motivates the importance of scale and region matching in few-shot learning.}
    \label{fig:scale_mismatch}
\end{figure*}

Though multi-scale modeling has been used in low-level vision problems, and matching features between regions is one of the oldest recognition tools  \cite{lazebnik_spmk,yang_sparse},  relation-based few-shot learning  (similarity learning between pairs of images) has not used such a mechanism despite clear benefits.

In addition to our matching mechanism, we embed the self-supervisory discriminators into our pipeline whose auxiliary task has to predict scale and spatial annotations, thus promoting a more discriminative training of encoder, attention and comparator. This is achieved by the use of Spatial-aware Discriminator (SD) to learn/predict the location and scale indexes of given features. Such strategies have not been investigated in matching, but they are similar to pretext tasks in self-supervised learning.

Beyond using SM on classic few-shot learning pipelines, we also propose a novel transformer-based pipeline, Spatially-aware Matching CrossTransformer (SmCT), which  learns the object correlations over locations and scales via cross-attention. Such a pipeline is  effective when being pre-trained on large-scale datasets. 

Below we summarize our contributions: 

\renewcommand{\labelenumi}{\roman{enumi}.}
\begin{enumerate}[leftmargin=0.6cm]
    \item We propose a novel spatially-aware matching few-shot learning strategy, compatible with many existing few-shot learning pipelines. We form possible region- and scale-wise pairs, and we pass them through comparator whose scores are re-weighted according to  the matching score of region- and scale-wise pairs obtained from the Spatial Matching unit.
    \item We propose  self-supervisory scale-level pretext tasks using second-order representations and auxiliary label terms for locations/scales, \eg, scale index.
    \item We investigate various matching strategies, \ie, different formulations of the objective, the use of sparsity-inducing regularization on attention scores, and the use of a balancing term on weighted average scores.
    \item We propose a novel and effective transformer-based  cross-attention matching strategy for few-shot learning, which learns object matching in pairs of images according to their respective locations and scales.
\end{enumerate}

\section{Related Work}  
\label{sec:related}

\noindent{\textbf{One- and few-shot learning }} has been  studied  in  shallow  \cite{miller_one_example,Li9596,NIPS2004_2576,BartU05,fei2006one,lake_oneshot} and deep learning setting  \cite{koch2015siamese,vinyals2016matching,snell2017prototypical,finn2017model,snell2017prototypical,sung2017learning,garcia2017few,rusu2018meta,gidaris2018dynamic,zhang2019power,zhang2019few,wertheimer2019few,Kim_2019_CVPR,Gidaris_2019_CVPR,can,parn,saml,dn4,mlso,ni2021anf}. 
Early works \cite{fei2006one,lake_oneshot}  employ generative models. 
Siamese Network \cite{koch2015siamese} is a two-stream CNN which can compare two streams. 

Matching Network \cite{vinyals2016matching} introduces the concept of support set and $L$-way $Z$-shot learning protocols to capture the similarity between a query and several support images in the episodic setting which we adopt. 
Prototypical Net~\cite{snell2017prototypical} computes distances between a query and prototypes of each class. %
Model-Agnostic Meta-Learning (MAML) \cite{finn2017model} introduces a meta-learning model which can be considered a form of transfer learning. 
Such a model was extended to Gradient Modulating MAML (ModGrad) \cite{christian_modgrad} to speed up the convergence. 
Relation Net~\cite{sung2017learning} 
 learns the relationship between query and support images 
 by a deep comparator that produces relation scores. 
SoSN \cite{zhang2019power} extends Relation Net \cite{sung2017learning} by second-order pooling. 
SalNet \cite{zhang2019few} is a saliency-guided end-to-end sample hallucinating model.  
Graph Neural Networks (GNN) \cite{uai_ke,ssgc_hao,coles_hao,rectifier_net,costa_kdd} can also be combined with few-shot learning  \cite{garcia2017few,Kim_2019_CVPR,Gidaris_2019_CVPR,jeanie,jeanie2,jeanie3}. In CAN \cite{can}, PARN \cite{parn} and RENet \cite{RENet},  self-correlation and cross-attention are employed to boost the performance. In contrast, our work studies explicitly matching over multiple scales and locations of input patches instead of features. Moreover, our SM is the first work studying how to combine self-supervision with spatial-matching to boost the performance. SAML \cite{saml} relies on a relation matrix to improve  metric measurements between local region pairs. DN4 \cite{dn4} proposes the deep nearest neighbor neural network to improve the image-to-class measure via deep local descriptors. Few-shot learning can also be performed in the transductive setting \cite{hao_fsl} and applied to non-standard problems, \eg, keypoint recognition \cite{keypoint_fsl}.

\paragraph{Second-order pooling }
has been used in texture recognition \cite{tuzel_rc} by  Region Covariance Descriptors (RCD),  in tracking \cite{porikli2006tracker} and object category recognition \cite{koniusz2017higher,koniusz2018deeper}. Higher-order statistics have  been used for action classification  \cite{koniusz2016tensor,tensor_act_tpami}, domain adaptation \cite{koniusz2017domain,yusuf_da}, few-shot learning \cite{zhang2019power,zhang2019few,wertheimer2019few,Zhang_2021_CVPR,maxexp},  few-shot object detection \cite{shan_fsl,xin6dof,Zhang_2022_CVPR,Zhang_eccv_2022} and even manifold-based incremental learning \cite{incremental}. 

We employ second-order pooling due to its (i)  permutation invariance (the ability to factor out spatial locations of feature vectors) and (ii) ability to capture second-order statistics. 

\begin{figure*}[t]
	\centering
	\includegraphics[width=\linewidth]{images/sm.pdf}
	\caption{\small The pipeline of \emph{SoSN+SM}. 
	We downsample input images twice (3 scales) and extract 5 sub-regions from the original image. Thus, 8 sub-images are passed through our encoder, and intermediate feature vectors are obtained and aggregated with SoP into matrices (red and blue blocks). We obtain $5$+$5$ location-wise support/query matrices $\mPsi^{i}_k$ and $\mPsi^{i}_q$ per support/query images $k$ and $q$, where $i\!\in\!\{1,\cdots,5\}$. We also obtain $3$+$3$ scale-wise matrices $\mPsi'^{i}_k$ and $\mPsi'^{i}_q$, where $i\!\in\!\{1,\cdots,3\}$. We pair them via relation operator $\vartheta$ (\eg, concatenation) into $25$ and $9$ relation descriptors passed to the attention mechanism and relation network, which produces weight scores $w_{pp'}$ ($5\!\times\!5$) and $w'_{ss'}$ ($3\!\times\!3$), and relation scores $\zeta_{pp'}$ ($5\!\times\!5$) and $\zeta'_{ss'}$ ($3\!\times\!3$), respectively. Finally, relation scores are re-weighted by attention scores and aggregated into the final score.
	}
	\label{fig:ms-fsl}
\end{figure*}

\paragraph{Notations.} 
Let $\vx\in\mbr{d}$ be a $d$-dimensional feature vector. $\idx{N}$ stands for the index set $\set{1, 2,\cdots,N}$. 
%
Capitalized boldface symbols such as $\mPhi$ denote matrices. Lowercase boldface symbols such as $\vphi$ denote vectors. Regular fonts such as $\Phi_{ij}$, $\phi_{i}$, $n$ or $Z$ denote scalars, \eg, $\Phi_{ij}$ is the $(i,j)$\textsuperscript{th} coefficient of $\mPhi$. Finally, $\delta(\vx\!-\!\vy)\!=\!1$ if $\vx\!=\!\vy$ and 0 otherwise.


\section{Approach}
Although spatially-aware representations have been studied in low-level vision, \eg,  deblurring, they have not been studied in relation-based learning (few-shot learning). Thus, it is not obvious how to match feature sets formed from pairs of images at different locations/resolutions. 

In conventional image classification, high-resolution images are known to be more informative than their low-resolution counterparts. However, extracting the discriminative information 
depends on the most expressive scale which varies between images. When learning to compare pairs of images (the main mechanism of relation-based few-shot learning), one has to match correctly same/related objects represented at two different locations and/or scales. 

Inspired by such issues, we show the importance of spatially-aware matching across locations and scales. To this end, we investigate our strategy on classic few-shot learning pipelines such as Prototypical Net, Relation Net and Second-order Similarity Network which we refer to as ({\em PN+SM}), ({\em RN+SM}) and ({\em SoSN+SM}) when combined with our Spatially-aware Matching (SM) mechanism. 
\subsection{Spatially-aware Few-shot Learning}
\label{sec:pipe}
Below, we take the {\em SoSN+SM} pipeline as an example to illustrate (Figure~\ref{fig:ms-fsl}) how we apply SM on the SoSN few-shot learning pipeline. 
We firstly generate spatially-aware image sequences from each original support/query sample, and feed them into the pipeline. Each image sequence includes 8 images, \ie, 5 location-wise crops and 3 scale-wise instances. Matching over such support-query sequences requires computing correlations between $8\!\times\!8=64$ pairs, which leads to significant training overhead when the model is trained with a large batch size. Thus, we decouple  spatially-aware matching into location-wise and scale-wise matching steps to reduce the computational cost. 

Specifically, let $\mathbf{I}^{1},\cdots,\mathbf{I}^{4}$ be four corners cropped from $\mathbf{I}$ of $84\times 84$ size without overlap and $\mathbf{I}^{5}$ be a center crop of $\mathbf{I}$. We refer to such an image sequence by $\{\mathbf{I}^p\}_{p\in\idx{5}}$, and they are of $42\!\times\!42$ resolution. Let $\mathbf{I}'^1$ be equal to the input image $\mathbf{I}$ of $84\times 84$ size, and $\mathbf{I}'^2$ and $\mathbf{I}'^3$ be formed by downsampling $\mathbf{I}$ to resolutions $42\!\times\!42$ and $21\!\times\!21$. We refer to these images by $\{\mathbf{I}'^s\}_{s\in\idx{3}}$. 
We pass these images via the encoding network $f(\cdot)$: 

\begin{equation}
    \mPhi^p = f(\mathbf{I}^p; \tF) \quad \!\text{and}\! \quad \mPhi'^s = f(\mathbf{I}'^s; \tF).
\end{equation}
where $\tF$ denotes parameters of encoder network, $\mPhi^p$ and $\mPhi'^s$ are  feature maps at the location $p$ and scale $s$, respectively.  As feature maps vary in size, we apply second-order pooling from Eq. \eqref{eq:sop} to these maps. We treat the channel mode as $D$-dimensional vectors and spatial modes $H\!\times\!W$ as $HW$ such vectors. As $N\!=\!HW$ varies, we define it to be $N$ for crops and $N'^s$ for scales. 
Then we form
\begin{equation}
     \mPsi^p=\eta\Big(\frac{1}{N}\mPhi^p\mPhi^{pT}\Big) \quad\! \text{and}\! \quad \mPsi'^s=\eta\Big(\frac{1}{N'^s}\mPhi'^s\mPhi'^{sT}\Big) .
    \label{eq:sop}
\end{equation}

Subsequently, we pass the location- and scale-wise second-order descriptors $\mPsi^p$ and $\mPsi'^s$ to the relation network (comparator) to model image relations. 
For the $L$-way $1$-shot problem, we have a support image ($k$\textsuperscript{th} index) with its image descriptors $(\mPhi_k^{p}, {\mPhi'}_k^{s})$ and a query image ($q$\textsuperscript{th} index) with its image descriptors  $(\mPhi_q^{p},{\mPhi'}_q^{s})$.
Moreover, each of the above descriptors belong to one of $L$ classes in the subset $\{c_1,\cdots,c_L\}\!\subset\!\idx{C}$ that forms the so-called $L$-way learning problem and the class subset $\{c_1,\cdots,c_L\}$ is chosen at random from $\idx{C}\!\equiv\!\{1,\cdots,C\}$. Then, the $L$-way $1$-shot relation requires relation scores:
\begin{equation}
\zeta^{pp'}_{kq}\!=\!r\left(\vartheta\!\left(\mPsi^p_k,\mPsi^p_q\!\right); \tR\right)
 \quad\!\!\text{and}\!\!\quad 
 \zeta'^{ss'}_{kq}\!=\!r\left(\vartheta\!\left(\mPsi'^s_k,\mPsi'^s_q\!\right); \tR\right),
 \label{eq:att}
\end{equation}
where $\zeta$ and $\zeta'$ are  relation scores for a $(k,q)$ image pair at locations $(p,p')$ and scales $(s,s')$. Moreover, $r(\cdot)$ is the relation network (comparator),  $\tR$ are its trainable parameters, $\vartheta(\cdot,\cdot)$ is the relation operator (we use concatenation along the channel mode). For $Z$-shot learning, this operator averages over $Z$ second-order matrices representing the support image before concatenating with the query matrix.


A naive loss for the location- \& scale-wise model is given as:
{\small
\begin{equation}
\!\mathit{L}=\sum\limits_{k,q,p}\! w_p\!\left(\zeta^{pp}_{kq}\!-\!\delta\!\left(l_k\!-\!l_q\right)\right)^2\!+\!\lambda\!\sum\limits_{k,q,s}\!w'_s\!\left(\zeta'^{ss}_{kq}\!-\!\delta\!\left(l_k\!-\!l_q\right)\right)^2\!,
\label{eq:naive1}
\end{equation}
}

\noindent{where} $l_k$ and $l_q$ refer to labels for support and query samples, $w_p$ and $w'_s$ are some priors (weight preferences) \wrt~locations $p$ and scales $s$. For instance, $w_p\!=\!1$ if $p\!=\!5$ (center crop),  $w_p\!=\!0.5$ otherwise, and $w_s\!=\!1/2^{s\!-\!1}$.

A less naive formulation assumes a  modified loss which performs matching between various regions and scales, defined as:
{\small
\begin{align}
\mathit{L}=\;&\sum\limits_{k,q}\sum\limits_{p,p'} \big(w_{pp'}^{kq}\big)^\gamma\!\big(\zeta^{pp'}_{kq}-\delta\!\left(l_k-l_q\right)\!\big)^2 
+\lambda\!\sum\limits_{k,q}\sum\limits_{s,s'} \big(w'^{kq}_{ss'}\big)^\gamma\!\big(\zeta'^{ss'}_{kq}\!-\!\delta\!\left(l_k\!-\!l_q\right)\!\big)^2,
\label{eq:naive2}
\end{align}
}

\noindent{where} $w_{pp'}$ and $w'_{ss'}$ are some pair-wise priors (weight preferences) \wrt~locations $(p,p')$ and scales $(s,s')$. We favor this formulation and we strive to learn $w_{pp'}^{kq}$ and $w'^{kq}_{ss'}$ rather than just specify rigid priors for all $(k,q)$ support-query pairs. Finally, coefficient $0\!\leq\!\gamma\!\leq\!\infty$ balances the impact of re-weighting. If $\gamma\!=\!0$, all weights are equal one. If $\gamma\!=\!0.5$, lower weights  contribute in a balanced way. If $\gamma\!=\!1$, we obtain regular re-weighting. If $\gamma\!=\!\infty$, the largest weight wins.

\begin{figure*}[t]
    \centering
    \includegraphics[width=\linewidth]{images/smct.pdf}
    \caption{Our Spatially-aware Matching CrossTransformer (SMCT) is built upon the cross-transformer \cite{ctx}. We introduce the spatially-aware  (location- and scale-wise) image sequences as inputs to exploit  cross-attention during feature matching.}
    \label{fig:sact}
\end{figure*}

\paragraph{Spatially-aware Matching.} As our feature encoder processes images at different scales and locations, the model should have the ability to select the best matching locations and scales for each support-query pair. Thus, we propose a pair-wise attention mechanism to re-weight (activate/deactivate) different matches withing support-query pairs when aggregating the final scores of comparator. Figure \ref{fig:ms-fsl} shows this principle.

Specifically, as different visual concepts may be expressed by their constituent parts (mixture of objects, mixture of object parts, \etc), each appearing at a different location or scale, we perform a soft-attention which selects $w_{pp'}\!\geq\!0$ and $w'_{ss'}\!\geq\!0$. Moreover, as co-occurrence representations are used as inputs to the attention network, the network selects a mixture of dominant scales and locations for co-occurring features (which may correspond to pairs of object parts). 

The Spatially-aware Matching (SM) network (two convolutional blocks and an FC layer) performs the location- and scale-wise matching respectively using shared model parameters. We opt for  a decoupled matching in order to reduce the training overhead. We perform $5\!\times\!5\!+\!3\!\times\!3\!=\!34$ matches per support-query pair rather than $5\!\times\!5\!\times\!3\!\times\!3\!=\!225$ matches but  a full matching variant is plausible (and could perform better). 
We have: 
\begin{equation}
\!\!\!\!w_{pp'}^{kq}\!=\!m\left(\vartheta(\!\cdot\!\mPsi^p_k, \mPsi^{p'}_q\!);\tM\right) \quad\text{and}\quad w'^{kq}_{ss'}\!=\!m\left(\vartheta(\!\cdot\!\mPsi'^s_k, \mPsi'^{s'}_q\!);\tM\right),
%
\end{equation}

\noindent{where} $m(\cdot)$ is the Spatially-aware Matching network, $\tM$ denotes its parameters, $(k,q)$ are query-support sample indexes. We impose a penalty to control the sparsity of matching:
\begin{align}
    &\Omega\!=\!\sum\limits_{k,q}\sum\limits_{p,p'}\Big\lvert w^{kq}_{pp'}\Big\lvert+\sum\limits_{k,q}\sum\limits_{s,s'}\Big\lvert w'^{kq}_{ss'}\Big\lvert.
\end{align}

Our spatially-aware matching network differs from the feature-based attention mechanism as we score the match between pairs of cropped/resized regions (not individual regions) to produce the attention map. 

\subsection{Self-supervised Scale and Scale Discrepancy}
\subsubsection{Scale Discriminator.} 
\begin{figure}[h]
	\centering
	\includegraphics[width=0.6\linewidth]{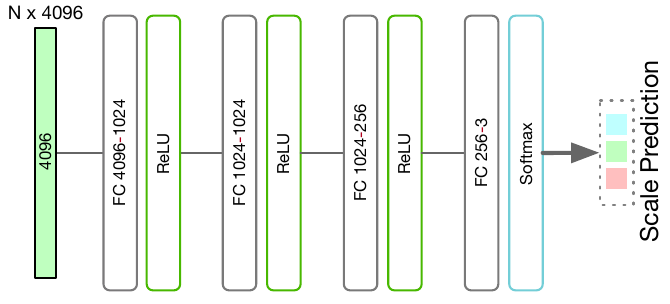}
	\caption{\small Scale Discriminator with 3 fully-connected layers.}
	\label{fig:scale_discriminator}
\end{figure}

To improve discriminative multi-scale representations, we employ self-supervision. We design a MLP-based Scale Discriminator (SD) as shown in Figure \ref{fig:scale_discriminator} which recognizes the scales of training images. 

Specifically, we feed second-order representations to the SD module and assign labels 1, 2 or 3 for $256\!\times\!256$, $128\!\times\!128$ or $64\!\times\!64$ images, respectively. We apply cross-entropy loss to train the SD module and classify the scale corresponding to given second-order feature matrix.

Given $\mPsi_i^s$ which is the second-order representation of $\mX_i^s$, we  vectorize them via $(:)$ 
and forward to the SD module to predict the scale index $s$. We have:
\begin{align}
    \mathbf{p}_i^s = \text{sd}(\mPsi_{i(:)}^s; \tC),
\end{align}
where $\text{sd}(\cdot)$ refers to the scale discriminator, $\tC$ denotes parameters of $\text{sd}(\cdot)$, and $\mathbf{p}$ are the scale prediction scores for $\mPsi$. 
We go over all $i$ corresponding to support and query images in the mini-batch and we use cross-entropy  to learn the parameters of Scale Discriminator: 
\begin{align}
    L_{sd} = -\sum\limits_{i,s} \text{log}\left(\frac{\exp({\mathbf{p}_i^s[s]})}{\sum_{s'}\exp({\mathbf{p}_{i}^{s}[s']})}\right),
    \label{eq:sss1}
\end{align}
where $s,s'\!\in\!\idx{S}$ enumerate over scale indexes.

\paragraph{Discrepancy Discriminator.} 
As relation learning requires comparing pairs of images, we propose to model scale discrepancy between each support-query pair by assign a discrepancy label to each pair. Specifically, we assign label $\Delta_{s,s^*}\!=\!s\!-\!s^*\!+\!1$ where $s$ and $s^*$ denote the scales of a given support-query pair. Then we train so-called  Discrepancy Discriminator (DD) to recognize the discrepancy between scales. DD uses the same architecture as SD while the input dimension is doubled due to concatenated support-query pairs on input. Thus:
\begin{align}
    \mathbf{p}_{ij}^{s,s*} = \text{dd}(\vartheta(\mPsi_i^s, \mPsi_j^{s*}); \tD),
\end{align}
where $\text{dd}(\cdot)$ refers to scale discrepancy discriminator, $\tD$ are the parameters of $dd$, $\mathbf{p}_{ij}^{s,s*}$ are  scale discrepancy prediction scores, $\vartheta$ is concat. in mode 3. We go over all $i,j$ support+query image indexes in the mini-batch and 
we apply the cross-entropy loss to learn the discrepancy labels: 
\begin{align}
\!\!\!\!L_{dd} = -\sum\limits_{i,s}\sum\limits_{j,s^*} \text{log}\left(\frac{\exp({\mathbf{p}_{ij}^{s,s^*}[\Delta_{s,s^*}]})}{\sum_{s'}\sum_{s'^*}\exp({\mathbf{p}_{ij}^{s,s^*}\![\Delta_{s',s'^*}]})}\right),
\end{align}
where where $s,s',s^*,s'^*\!\in\!\idx{S}$ enumerate over scale indexes.

\paragraph{Final Loss.}  
The total loss combines the proposed Scale Selector, Scale Discriminator and Discrepancy Discriminator:
\begin{align}
    \argmin\limits_{\tF, \tR, \tM, \tC, \tD} \quad \alpha \Omega + L + \beta L_{sd} + \gamma L_{dd},
\end{align}
where $\alpha, \beta, \gamma$ are the hyper-parameters that control the impact of the regularization and each individual loss component.

\subsection{Transformer-based Spatially-aware Pipeline}
Our SM network can be viewed as an instance of attention, whose role is to re-weight numbers of spatial pairs to improve the discriminative relation learning between support and query samples. Recently, transformers have proven  very effective in learning the discriminative representations in both natural language processing and computer vision tasks. Inspired by the self-attention \cite{ctx}, which can naturally be used to address the feature matching problem, we further develop a novel Spatially-aware Matching CrossTransformer (SmCT) to match location- and scale-wise support-query patches in few-shot learning. Figure \ref{fig:sact} shows the architecture of SmCT, which consists of 4 heads, namely the support key and value heads, and the query key and value heads. In contrast to using SM in classic pipelines, where we decoupled the matching into location-wise and scale-wise steps, all possible matching combinations are considered in the SmCT pipeline. Thus, we do not use symbols $\mathbf{I}^p$ and $\mathbf{I}'^s$ in this section. The  image sequence is simply given by $\{\mathbf{I}^s\}_{s\in\idx{8}}$.

\begin{table*}[t]
\centering
\caption{\small Evaluations on  \textit{mini}-ImageNet and \textit{tiered}-ImageNet  (5-way acc. given). }
\label{table_mini}
\makebox[\textwidth]{
\setlength{\tabcolsep}{0.4em}
\fontsize{8}{9}\selectfont
\hspace{-0.3cm}
\begin{tabular}{llccccc}
\toprule
& & & \multicolumn{2}{c}{\textit{mini}-ImageNet} & \multicolumn{2}{c}{\textit{tiered}-ImageNet} \\ 
Model & \kern-6.0em& \kern-0.3em Backbone & 1-shot & 5-shot & 1-shot & 5-shot \\ \hline
\textit{MN} & \kern-1.3em\cite{vinyals2016matching}\kern-4.2em &\kern-0.3em - & $43.56 \pm 0.84 $ & $55.31 \pm 0.73 $ & - & - \\
\textit{PN} &\kern-1.3em\cite{snell2017prototypical}\kern-3.7em &\kern-0.3em Conv--4--64 & $49.42 \pm 0.78 $ & $68.20 \pm 0.66$ & $53.31 \pm 0.89$ & $72.69 \pm 0.74$\\
\textit{MAML} &\kern-1.3em\cite{finn2017model}\kern-3.7em &\kern-0.3em Conv--4--64 & $48.70 \pm 1.84 $ & $63.11 \pm 0.92 $ & $51.67 \pm 1.81$ & $70.30 \pm 1.75$\\ 
\textit{RN} &\kern-1.3em\cite{sung2017learning}\kern-3.7em &\kern-0.3em Conv--4--64 & $50.44 \pm 0.82 $ & $65.32 \pm 0.70 $  & $54.48\pm0.93$ & $71.32\pm0.78$\\
\textit{GNN} &\kern-1.3em\cite{garcia2017few}\kern-3.7em&\kern-0.3em Conv--4--64 & $50.30 $ & $66.40 $ & - & - \\
\textit{MAML++} &\kern-1.3em\cite{maml++}\kern-3.7em&\kern-0.3em Conv--4--64 & $52.15 \pm 0.26 $ & $68.32 \pm 0.44$ & - & -\\
\textit{SalNet} &\kern-1.3em\cite{zhang2019few}\kern-3.7em&\kern-0.3em Conv--4--64 & $57.45 \pm 0.86 $ & $72.01 \pm 0.75 $  & - & -\\
\textit{SoSN} &\kern-1.3em\cite{zhang2019power}\kern-3.7em&\kern-0.3em Conv--4--64 & $52.96 \pm 0.83 $ & $68.63 \pm 0.68 $  & - & - \\
\textit{TADAM} &\kern-1.3em\cite{tadam} & ResNet-12 & $58.50 \pm 0.30$ & $76.70 \pm 0.30$ & - & - \\
\textit{MetaOpt}&\kern-1.3em\cite{metaopt}\kern-3.7em &\kern-0.3em ResNet-12 & $61.41 \pm 0.61 $ & $77.88 \pm 0.46 $ & $65.99\pm0.72$ & $81.56\pm0.53$\\
\textit{DeepEMD}&\kern-1.3em\cite{zhang2020deepemd}\kern-3.7em & ResNet-12 & $65.91 \pm 0.82 $ & $82.41 \pm 0.56$  & - & - \\
\hline
\rowcolor{LightCyan}\textit{PN+SM}\kern-1.0em &\kern-6.0em &\kern-0.3em Conv--4--64 & ${52.01\pm 0.80}$ & ${69.92 \pm 0.67}$ & $54.37\pm0.82$ & $75.13\pm0.77$\\
\rowcolor{LightCyan}\textit{RN+SM}\kern-1.0em &\kern-6.0em  &\kern-0.3em Conv--4--64 & ${54.99 \pm 0.87}$ & ${68.57 \pm 0.63}$ & $57.01\pm0.91$ & $75.04\pm0.78$\\
\rowcolor{LightCyan}\textit{SoSN+SM}\kern-1.0em &\kern-6.0em  &\kern-0.3em Conv--4--64 & ${57.11 \pm 0.84}$ & ${71.98\pm 0.63}$ & $61.58\pm0.90$ & $78.64\pm0.75$\\
\rowcolor{LightCyan}\textit{SoSN+SM}\kern-1.0em &\kern-6.0em  &\kern-0.3em ResNet-12 & ${62.36\pm 0.85}$ & ${78.86 \pm 0.63}$ & $66.35\pm0.91$ & $82.21\pm0.71$ \\
\rowcolor{LightCyan}\textit{DeepEMD+SM}\kern-0.4em &\kern-6.0em  &\kern-0.3em ResNet-12 & $\mathbf{66.93\pm 0.84}$ & $\mathbf{84.34 \pm 0.61}$ & $\mathbf{70.19\pm0.89}$ & $\mathbf{86.98\pm0.74}$ \\
\bottomrule
\end{tabular}}
\end{table*}

\begin{table*}[t]
\centering
\caption{Evaluations on the Open MIC dataset (Protocol I) (1-shot learning accuracy). ({\fontsize{9}{9}\selectfont \url{http://users.cecs.anu.edu.au/~koniusz/openmic-dataset}}).$\!\!\!\!\!$}
\label{table_openmic}
\makebox[\textwidth]{
\setlength{\tabcolsep}{0.1em}
\renewcommand{\arraystretch}{1}
\fontsize{6}{6}\selectfont
\begin{tabular}{l|c|c|c|c|c|c|c|c|c|c|c|c|c}
\hline
Model & way & $p1\!\!\rightarrow\!\!p2$ & $p1\!\!\rightarrow\!\!p3$& $p1\!\!\rightarrow\!\!p4$& $p2\!\!\rightarrow\!\!p1$& $p2\!\!\rightarrow\!\!p3$ &$p2\!\!\rightarrow\!\!p4$& $p3\!\!\rightarrow\!\!p1$& $p3\!\!\rightarrow\!\!p2$& $p3\!\!\rightarrow\!\!p4$& $p4\!\!\rightarrow\!\!p1$& $p4\!\!\rightarrow\!\!p2$& $p4\!\!\rightarrow\!\!p3$\\
\hline
\textit{RN} \cite{sung2017learning}  &\multirow{4}{*}{5}& $71.1$ & $53.6$ & $63.5$ & $47.2$ & $50.6$ & $68.5$ & $48.5$ & $49.7$ & $68.4$ & $45.5$ & $70.3$ & $50.8$\\
\textit{SoSN$_{(84)}$} \cite{zhang2019power} &&  $81.4$ & ${65.2}$ & ${75.1}$ & ${60.3}$ & ${62.1}$ & ${77.7}$ & ${61.5}$ & ${82.0}$ & ${78.0}$ & ${59.0}$ & ${80.8}$ & ${62.5}$\\
\it SoSN$_{(256)}$ &&  $84.1$ & $69.3$ & $82.5$ & $64.9$ & $66.9$ & $82.8$ & ${65.8}$ & ${85.1}$ & ${81.1}$ & ${65.1}$ & ${83.9}$ & ${66.6}$\\
\rowcolor{LightCyan}\it SoSN+SM  &&  $\mathbf{85.6}$ & $\mathbf{73.6}$ & $\mathbf{85.0}$ & $\mathbf{67.7}$ & $\mathbf{69.6}$ & $\mathbf{83.1}$ & $\mathbf{68.2}$ & $\mathbf{86.9}$ & $\mathbf{82.9}$ & $\mathbf{67.4}$ & $\mathbf{84.7}$ & $\mathbf{68.4}$\\
\hline
\textit{RN} \cite{sung2017learning}&\multirow{4}{*}{20}& $40.1$ & $30.4$ & $41.4$ & $23.5$ & $26.4$ & $38.6$ & $26.2$ & $25.8$ & $46.3$ & $23.1$ & $43.3$ & $27.7$\\
\textit{SoSN$_{(84)}$} \cite{zhang2019power}&&   ${61.5}$ & ${42.5}$ & ${61.0}$ & ${36.1}$ & ${38.3}$ & ${56.3}$ & ${38.7}$ & ${59.9}$ & ${59.4}$ & ${37.4}$ & ${59.0}$ & ${38.6}$\\
\it SoSN$_{(256)}$ &&  ${63.9}$ & ${49.2}$ & ${65.9}$ & ${43.1}$ & ${44.6}$ & ${62.6}$ & ${44.2}$ & ${63.9}$ & ${64.1}$ & ${43.8}$ & ${63.1}$ & ${44.3}$\\
\rowcolor{LightCyan}\it SoSN+SM          &&  ${65.5}$ & ${51.1}$ & ${67.6}$ & ${45.2}$ & ${46.3}$ & ${64.5}$ & ${46.3}$ & ${66.2}$ & ${67.0}$ & ${45.3}$ & ${65.8}$ & ${47.1}$\\
\hline
\textit{RN} \cite{sung2017learning}&\multirow{4}{*}{30}& $37.8$ & $27.3$ & $39.8$ & $22.1$ & $24.3$ & $36.7$ & $24.5$ & $23.7$ & $44.2$ & $21.4$ & $41.5$ & $25.5$\\
\textit{SoSN$_{(84)}$} \cite{zhang2019power}&&  $60.6$ &  $40.1$ & $58.3$ & $34.5$ & $35.1$ & $54.2$ & $36.8$ & $58.6$ & $56.6$ & $35.9$ & $57.1$ & $37.1$\\
\it SoSN$_{(256)}$ &&  $61.7$ & $46.6$ & $64.1$ & $41.4$ & $40.9$ & $60.3$ & $41.6$ & $ 61.0$ & $ 60.0$ & $ 42.4$ & $ 61.2$ & $ 41.4$\\
\rowcolor{LightCyan}\it SoSN+SM &&           $62.6$ & $47.3$ & $65.2$ & $41.9$ & $41.7$ & $61.5$ & $43.1$ & $ 61.8$ & $ 61.0$ & $ 43.1$ & $ 62.1$ & $ 42.3$\\
\hline
\end{tabular}}
\text{\small \quad\!\!p1: shn+hon+clv, p2: clk+gls+scl, p3: sci+nat, p4: shx+rlc.} 
\text{Notation {\em x$\rightarrow$y} means training on exhibition {\em x} and testing on {\em y}. }
\end{table*}

We pass the spatial sequences of support and query samples $\mathbf{I}^s_k$ and $\mathbf{I}^s_q$ into the backbone and obtain $\mPhi^s_k,\mPhi^{\tilde{s}}_q\!\in \!\mathbb{R}^{D\times N}$. Due to different scales  of feature maps for $s\!=\!6,7,8$, we downsample larger feature maps and obtain $N'\!=\!10\!\times\!10$ feature vectors. We feed them into the  key/value heads to obtain keys $\mathbf{K}^s_k,\mathbf{K}^{\tilde{s}}_q\!\in\!\mathbb{R}^{d_k \times N'}$  and values $\mathbf{V}^s_k, \mathbf{V}^{\tilde{s}}_q\!\in\!\mathbb{R}^{d_v \times N'}$. 
Then we multiply support-query spatial key pairs followed by  SoftMax in order to obtain the normalized cross-attention scores $\mathbf{C}_{s\tilde{s}}\!=\! \text{SoftMax}({\mathbf{K}^s_k}^T\mathbf{K}^{\tilde{s}}_q)\!\in\! \mathbb{R}^{N'\times N'}$, which are used as correlations in aggregation of support values \wrt  each location and scale. We obtain the aligned spatially-aware prototypes $\tilde{\mathbf{V}}^s_k\!\in\!\mathbb{R}^{d_v \times N'}$:
\begin{align}
    \tilde{\mathbf{V}}^s_k\!=\!\sum\limits_{\tilde{s}}\mathbf{V}^s_k \mathbf{C}_{s\tilde{s}}.
\end{align}
We measure  Euclidean distances between the aligned prototypes and corresponding query values, which act as the final similarity between sample $\mathbf{I}_k$ and $\mathbf{I}_q$: 
\begin{align}
&    L\!=\!\sum\limits_{k,q} \big(\zeta_{kq} \!-\! \delta(l_k\!-\!l_q)\big)^2
\quad\text{where}\quad \zeta_{kq}\!=\!\sum\limits_s \parallel \tilde{\mathbf{V}}^s_k\!-\!\tilde{\mathbf{V}}^s_q \parallel_F^2.
\end{align}

\begin{table*}[t]
    \centering
    \caption{\small The experiments on selected subsets of Meta-Dataset (train-on-ILSVRC setting). We compare our Spatially-aware Matching pipelines with recent baseline models. Following the training steps in \cite{ctx}, we also apply the SimCLR episodes to train our SmCT (which uses the ResNet-34 backbone).}
    \makebox[\textwidth]{
    \setlength{\tabcolsep}{0.8em}
    \fontsize{9}{9}\selectfont
    \begin{tabular}{llcccccc}
    \toprule
    &\kern-2.5em  &  ImageNet & Aircraft & Bird & DTD & Flower & Avg  \\ \hline
    \textit{k-NN} &\kern-3.5em\cite{metadataset}\kern-4.5em & 41.03 & 46.81 & 50.13 & 66.36 & 83.10 & 57.49\\
    \textit{MN} &\kern-3.5em\cite{vinyals2016matching}\kern-4.5em & 45.00 & 48.79 & 62.21 & 64.15 & 80.13 & 60.06 \\
    \textit{PN} &\kern-3.5em\cite{snell2017prototypical}\kern-4.5em & 50.50 & 53.10 & 68.79 & 66.56 & 85.27 & 64.84 \\
    \textit{RN} &\kern-3.5em\cite{sung2017learning}\kern-4.5em& 34.69 & 40.73 & 49.51 & 52.97 & 68.76 & 49.33 \\
    \textit{SoSN} &\kern-3.5em\cite{zhang2019power}\kern-4.5em & 50.67 & 54.13 & 69.02 & 66.49 & 87.21 & 65.50 \\
    \textit{CTX} &\kern-3.5em\cite{ctx}\kern-4.5em & 62.76 & 79.49 & 80.63 & 75.57 & 95.34 & 78.76 \\ \hline
    \rowcolor{LightCyan}\textit{PN+SM} &\kern-5em & 53.12 & 57.06 & 72.01 & 70.23 & 88.96 & 68.28 \\
    \rowcolor{LightCyan}\textit{RN+SM} &\kern-5em & 41.07 & 46.03 & 53.24 & 58.01 & 72.98 & 54.27 \\
    \rowcolor{LightCyan}\textit{SoSN+SM} &\kern-5em & 52.03 & 56.68 & 70.89 & 69.03 & 90.36 & 67.80 \\
    \rowcolor{LightCyan}\textit{SMCT} &\kern-5em & \textbf{64.12} & \textbf{81.03} & \textbf{82.98} & \textbf{76.95} & \textbf{96.45} & \textbf{80.31} \\
    \bottomrule
    \end{tabular}}
    \label{tab:meta-dataset}
\end{table*}

\begin{table}[t]
\centering
\caption{\small Evaluations on  fine-grained recognition datasets, Flower-102, CUB-200-2011 and Food-101 (5-way acc. given). See \cite{sung2017learning,zhang2019power} for details of baselines listed in this table.}
\label{table_fgc}
\makebox[\linewidth]{
\setlength{\tabcolsep}{1em}
\fontsize{9}{10}\selectfont
\begin{tabular}{lc cc cc c}
\toprule
& \multicolumn{2}{c}{Flower-102} & \multicolumn{2}{c}{CUB-200-2011} & \multicolumn{2}{c}{Food-101} \\
Model & 1-shot & 5-shot & 1-shot & 5-shot & 1-shot & 5-shot \\ \hline
\textit{PN}& $62.81$ & $82.11$ & $37.42$ & $51.57$ & $36.71$ & $53.43$  \\ 
\textit{RN} & $68.52$ & $81.11$ & $40.36$ & $54.21$ & $36.89$ & $49.07$  \\ 
\textit{SoSN}& $76.27$ & $88.55$ & $47.45$ & $63.53$ & $43.12$ & $58.13$  \\
\midrule
\rowcolor{LightCyan}\textit{RN+SM} & ${71.69}$ & ${84.45}$ & ${45.79}$ & ${58.67}$ & ${45.31}$ & ${55.67 }$\\
\rowcolor{LightCyan}\textit{SoSN+SM} & $\mathbf{81.69}$ & $\mathbf{91.21}$ & $\mathbf{54.24}$ & $\mathbf{70.85}$ & $\mathbf{48.86}$ & $\mathbf{63.67}$\\
\bottomrule
\end{tabular}}
\end{table}

\section{Experiments}
\label{sec:exp}
Below we demonstrate usefulness of our proposed Spatial- and Scale-matching Network by evaluations (one- and few-shot protocols) on \textit{mini}-ImageNet \cite{vinyals2016matching},  \textit{tiered}-ImageNet \cite{ren18fewshotssl}, Meta-Dataset \cite{metadataset} 
and fine-grained  datasets.

\paragraph{Setting.}  
We use the standard $84\!\times\!84$ image resolution for \textit{mini}-ImageNet and fine-grained datasets, and $224\!\times\!224$ resolution for Meta-Dataset for fair comparisons. Hyper-parameter $\alpha$ is set to 0.001 while $\beta$ is set to 0.1 via cross-validation on \textit{mini}-ImageNet. Note that SmCT uses ResNet-34 backbone on Meta-Dataset, and ResNet-12 on \textit{mini}-ImageNet and \textit{tiered}-ImageNet. The total number of training episode is 200000, and the number of testing episode is 1000.

\subsection{Datasets}
Below, we describe our experimental setup and datasets. 

\paragraph{\textit{mini}-ImageNet} \cite{vinyals2016matching} consists of 60000 RGB images from 100 classes.
We follow the standard protocol (64/16/20 classes for training/validation/testing). 

\paragraph{\textit{tiered}-ImageNet} \cite{ren18fewshotssl} consists of 608 classes from ImageNet. We follow the protocol that uses 351 base classes, 96 validation and 160 test classes.

\paragraph{Open MIC} is the Open Museum Identification Challenge (Open MIC) \cite{me_museum}, a dataset with photos of various museum exhibits, \eg, paintings, timepieces, sculptures, glassware, relics, science exhibits, natural history pieces, ceramics, pottery, tools and indigenous crafts, captured from 10 museum spaces according to which this dataset is divided into 10 subproblems. In total, it has 866 diverse classes and 1--20 images per class. We combine ({\em shn+hon+clv}), ({\em clk+gls+scl}), ({\em sci+nat}) and ({\em shx+rlc}) into subproblems {\em p1}, $\!\cdots$, {\em p4}. We form 12 possible pairs in which subproblem $x$ is used for training and $y$ for testing (x$\rightarrow$y).

\paragraph{Meta-Dataset} \cite{metadataset} is a recently proposed benchmark consisting of 10 publicly available datasets to measure the generalized performance of each model. The Train-on-ILSVRC setting means that the model is merely trained via ImageNet training data, and then evaluated on the test data of remaining 10 datasets. In this paper, we follow the Train-on-ILSVRC setting. We choose 5 datasets to measure the overall performance. 

\paragraph{Flower-102} \cite{Nilsback08} contains 102 fine-grained classes of  flowers. Each class has 40--258 images. We randomly select 80/22 classes for training/testing.

\paragraph{Caltech-UCSLD-Birds 200-2011 (CUB-200-2011)} \cite{WahCUB_200_2011} has 11788 images of 200 fine-grained bird species, 150 classes a for training and the rest for testing.

\paragraph{Food-101} \cite{bossard14} has 101000 images of 101 fine-grained classes, and 1000 images per category. We choose 80/21 classes for training/testing.

\begin{figure*}[t]
    \centering
    \includegraphics[width=\linewidth]{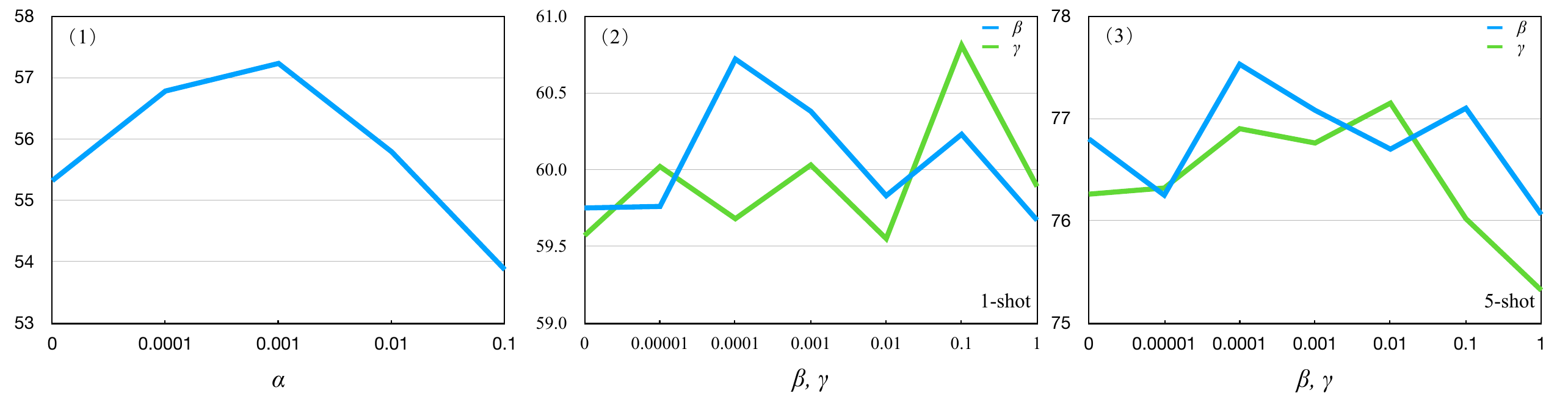}
    \caption{Accuracy \wrt $\alpha$ (subfigure 1), $\beta$ (the blue curve in subfigure 2\&3) and $\gamma$ (the red curve in subfigure 2\&3), which control  the impact of SD and DD.}
    \label{fig:hyper-pram}
\end{figure*}

\begin{figure*}[t]
    \centering
    \includegraphics[width=\linewidth]{images/sparsity.pdf}
    \caption{The histograms of spatial-matching scores with different $\alpha$ values, which demonstrate how $\Omega$ induces sparsity and improves the performance.}
    \label{fig:sparsity}
\end{figure*}

\begin{figure*}[t]
    \centering
    \includegraphics[trim=0 3 0 0, clip=true,width=\linewidth]{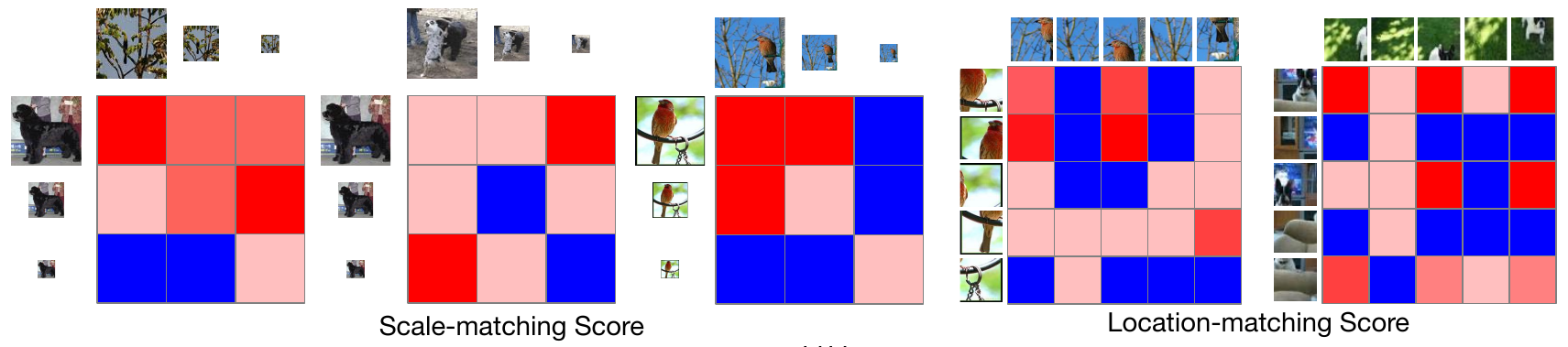}
    \caption{The histograms of spatial-matching scores with different $\alpha$ values, which demonstrate how $\Omega$ induces sparsity and improves the performance.}
    \label{fig:match-score}
\end{figure*}

\begin{table}[t]
\centering
\caption{\small Ablation studies \wrt~the location- and scale-wise inputs (no matching used). Region crops and scale selection is done either in the image space ({\em img}) or on the feature maps ({\em feat.}), (5-way 1-shot accuracy, ‘SoSN+SM’ with ResNet-12 backbone).}
\label{table_ablation}
\makebox[\linewidth]{
\setlength{\tabcolsep}{0.5em}
\fontsize{9}{10}\selectfont
\begin{tabular}{cccc}
\toprule
Scale-wise ({\em img}) & Scale-wise ({\em feat.}) & Loc.-wise ({\em img}) & Loc.-wise ({\em feat.}) \\ 
\midrule
59.51& 58.54 & 60.14 & 58.16 \\
\bottomrule
\end{tabular}}
\end{table}

\subsection{Performance Analysis}

Table \ref{table_mini} shows our evaluations on \textit{mini}-ImageNet \cite{vinyals2016matching} and \textit{tiered}-ImageNet. Our approach achieves the state-of-the-art performance among all methods based on both 'Conv-4-64' and 'ResNet-12' backbones at $84\!\times\!84$ input resolution. 
By adding SM (Eq. \eqref{eq:att}) and self-supervisory loss (Eq. \eqref{eq:sss1}), we achieve $57.11$ and $71.98$ scores for 1- and 5-shot protocols with the SoSN baseline (Conv-4-64 backbone), and $66.93$ and $85.34$ with the DeepEMD baseline (ResNet-12 backbone) on \textit{mini}-ImageNet. The average improvements gained from SM are 4\% and 3.5\% with the Conv-4-64 backbone, 1\% and 1.9\% with the ResNet-12 backbone for 1- and 5-shot protocols. These results outperform results in previous works, which strongly supports the benefit of our spatially-aware matching. On \textit{tiered}-ImageNet, our proposed model obtains $3\%$ and $4\%$ improvement for 1-shot and 5-shot protocols with both the Conv-4-64 and the ResNet-12 backbones. Our performance on \textit{tiered}-ImageNet  is also better than in previous works. 
However, we observed that SmCT with the ResNet-12 backbone does not perform as strongly on the above two datasets. A possible reason is that the complicated transformer architecture  overfits when the scale of training dataset is small (in contrast to its performance on Meta-Dataset). 

We also evaluate our proposed network on the Open MIC dataset \cite{me_museum}. Table~\ref{table_openmic} shows that our proposed method  performs better than baseline models. By adding SM and self-supervised discriminators into baseline models, the accuracies are improved by $1.5\% \sim 3.0\%$.
 
Table \ref{tab:meta-dataset} presents results on the Train-on-ISLVRC setting on 5 of 10 datasets of Meta-Dataset. Applying SM on classic simple baseline few-shot leaning methods brings impressive improvements on all datasets. Furthermore, our SmCT achieves the state-of-the-art results compared to previous methods. This observation is consistent with our analysis that transformer-based networks are likely to be powerful when being trained on a large-scale dataset. 

Table \ref{table_fgc} shows that applying SM on classic baseline models significantly outperforms others on fine-grained Flower-102, CUB-200-2011 and Food-101. 
 

\paragraph{Spatially-aware Matching.} The role of SM is to learn matching between regions and scales of support-query pairs. As shown in Table \ref{table_ablation}, using the Spatially-aware Matching Network can further improve results for both 1-shot and 5-shot learning on \textit{mini}-ImageNet by $1.5\%$ and $1.0\%$, respectively. The results on Flower-102, Food-101  yield $\sim\!1.5\%$ gain for SmN. Figure \ref{fig:hyper-pram} shows the impact of $\alpha$, $\beta$ and $\gamma$ on the performance. Figure \ref{fig:sparsity} verifies the usefulness of $\Omega$ regularization term. To this end, we show histograms of spatial-matching scores to demonstrate how $\alpha$ affects the results \vs~sparsity of matching scores. For instance, one can see the results are the best for moderate $\alpha\!=\!0.001$, and the bin containing null counts also appears larger compared to  $\alpha\!=\!0$ (desired behavior). Figure \ref{fig:match-score} visualizes the  matching scores. We randomly sample  support-query pairs to show that visually related patches have higher match scores (red) than unrelated pairs (blue). Do note the location- and scale- discrimination is driven by matching both similar and dissimilar support-query pairs driven by the relation loss.

\paragraph{Self-supervised Discriminators.} Pretext tasks are known for their ability to boost the performance of image classification due to additional regularization they provide. Applying Scale Discriminator (SD) and Discrepancy Discriminator (DD) is  an easy and cheap way to boost the representational power of our network. Pretext tasks do not affect the network complexity or training times. According to our evaluations, the SD improves the 1-shot accuracy by $1.1\%$ and 5-shot accuracy by $1.0\%$, while the DD improves the accuracy by $1.2\%$ and $1.1\%$ $0.9\%$ for 1- and 5-shot respectively on the \textit{mini}-ImageNet dataset.

In summary, without any pre-training, combining our Spatially-aware Matching strategy brings consistent improvements on the Conv-4-64 and ResNet-12 backbones and various few-shot learning methods, with the overall accuracy outperforming  state-of-the-art methods on all few-shot learning datasets. Our novel transformer-based  SmCT model  also performs strongly on the recently proposed Meta-Dataset, which further supports the usefulness of spatial modeling.

\section{Conclusions}
We have proposed the Spatially-aware Matching strategy for few-shot learning, which is shown to be orthogonal to the choice of baseline models and/or backbones. Our novel feature matching mechanism  helps models learn a more accurate similarity due to matching multiple locations and coarse-to-fine scales of support-query pairs. We show how to leverage a self-supervisory pretext task based on spatial labels. We have also propsoed a novel Spatially-aware Matching CrossTransformer to perform matching via the recent popular self- and cross-attention strategies. Our experiments demonstrate the usefulness of the proposed SM strategy and SmCT in capturing accurate image relations. Combing our SM with various baselines outperforms previous works in the same testbed. SmCT achieves SOTA results on large-scale training data.

\section*{Acknowledgements}
This work is supported by National Natural Science Fundation of China (No. 62106282), and Young Elite Scientists Sponsorship Program by CAST (No. 2021-JCJQ-QT-038). Code:  \url{https://github.com/HongguangZhang/smfsl-master}.
\bibliographystyle{splncs}
\bibliography{egbib}
\end{document}